\documentclass[journal,onecolumn]{IEEEtran}

\usepackage{amsmath,amsfonts}
\usepackage{algorithmic}
\usepackage{algorithm}
\usepackage{array}
\usepackage[caption=false,font=normalsize,labelfont=sf,textfont=sf]{subfig}
\usepackage{textcomp}
\usepackage{stfloats}
\usepackage{url}
\usepackage{verbatim}
\usepackage{graphicx}
\usepackage{cite}

\usepackage{amssymb}
\usepackage[dvipsnames]{xcolor}
\usepackage{tikz}
\usepackage{pgfplots} 
\usetikzlibrary{calc, through, backgrounds}

\usepackage{notationSIAMESE}

\begin{document}

\title{Forensic Study of Paintings Through the Comparison of Fabrics}

\author{Juan José Murillo-Fuentes, Pablo M. Olmos, Laura Alba-Carcelén\thanks{J.J. Murillo-Fuentes is with the Dep. Signal Processing and Communications, ETSi Universidad de Sevilla, Sevilla. Spain.}\thanks{Pablo M. Olmos is with the Dep. Signal Processing and Communications, Universidad Carlos III de Madrid, Leganés, Spain}\thanks{Laura Alba-Carcelén is with the Gabinete Técnico, Museo Nacional del Prado, Madrid, Spain}
\thanks{This work was supported by Ministerio de Ciencia e Innovación de España and FEDER-European Union [EPiCENTER, PID2021-123182OB-I00], [MCIN/AEI/10.13039/501100011033], [Skin, PID2021-127871OB-I00].}
\thanks{Manuscript received April 19, 2025; revised August 16, 2025.}}

\markboth{Submitted to IEEE Trans. Image Processing, Apr.~2025}%
{Shell \MakeLowercase{\textit{et al.}}: A Sample Article Using IEEEtran.cls for IEEE Journals}


\maketitle

\begin{abstract}
The study of canvas fabrics in works of art is a crucial tool for authentication, attribution and conservation. Traditional methods are based on thread density map matching, which cannot be applied when canvases do not come from contiguous positions on a roll. This paper presents a novel approach based on deep learning to assess the similarity of textiles. We introduce an automatic tool that evaluates the similarity between canvases without relying on thread density maps. A Siamese deep learning model is designed and trained to compare pairs of images by exploiting the feature representations learned from the scans. In addition, a similarity estimation method is proposed, aggregating predictions from multiple pairs of cloth samples to provide a robust similarity score. Our approach is applied to canvases from the Museo Nacional del Prado, corroborating the hypothesis that plain weave canvases, widely used in painting, can be effectively compared even when their thread densities are similar. The results demonstrate the feasibility and accuracy of the proposed method, opening new avenues for the analysis of masterpieces.

\end{abstract}

\section{Introduction}
\subsection{Significance of Fabric Analysis}

\IEEEPARstart{I}{n} the forensic examination of canvas masterpieces, the support analysis is of paramount importance~\cite{Alba21}. Conclusions regarding a set of masterpieces sharing a specific type of support are crucial in the historical study of artworks. Among painting supports, cloth is the most prevalent, and within cloth supports, the plain weave\footnote{Also known as taffeta, tabby, or calico weave.} offers an optimal balance between robustness and simplicity, making it a first choice among painters~\cite{Vanderlip80}.

\begin{figure}[htbp]
\centering
\begin{tabular}{cc}
\includegraphics[width=3.5cm]{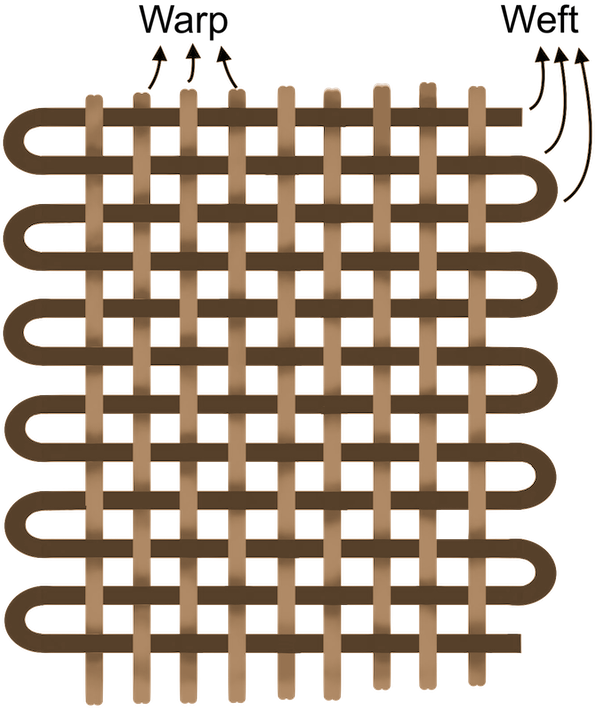} & \includegraphics[width=4.0cm]{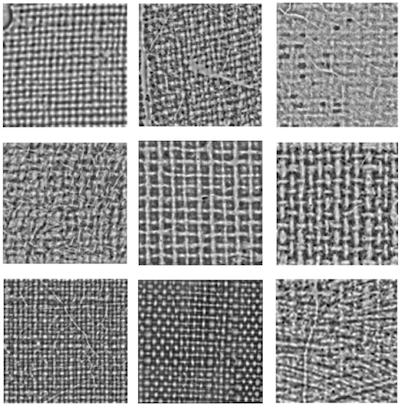}\\
(a) & (b)
\end{tabular}
\caption{(a) A sketch of plain weave with warp and weft yarns in weaving as described in~\cite{Barlow1878}, and (b) patches of 1 cm-sided X-ray plates from different paintings using plain weave fabrics.}
\label{fig:Taffeta}
\end{figure}

The plain weave is a simple structure comprising vertical and horizontal threads interlaced, as depicted in Figure~\ref{fig:Taffeta}(a). In a loom, a set of threads known as the warp is arranged in parallel. Subsequently, the weaver passes another thread orthogonally to the warp from one side to the other, tightens it, and passes it back, repeating the process. This orthogonal set of threads is called the weft. The pattern of the plain weave has been extensively studied, particularly in the frequency domain~\cite{Escofet2001, Johnson2013, Simois18}. Although this pattern is common to all plain weave fabrics, significant differences can be observed from one canvas to another:

\begin{itemize}
\item The separation of warp, or weft, threads might vary, with mean thread densities typically ranging from 5 to 25 threads per centimeter. Moreover, significant warp or weft thread density variations may be found within the same roll.


\item The type of material used for the threads may vary from one fabric to another, such as linen, cotton, silk, etc.

\item The thickness of the threads may differ along the warp, the weft, or between warp and weft.

\item A greater tension applied to the warp makes the weft less noticeable. The same applies if the weft threads have larger tension in the loom.
\end{itemize}

Due to these variations, it has been found in practice that canvases with similar average thread counts can exhibit noticeably different textures \cite{Simois18}. 

%
%
 %
%
In fabric analysis using image processing and computer vision, the X-ray image of the canvas is commonly utilized, as usually the original fabric cannot be directly observed due to relining performed to protect and reinforce it. The intensity of each pixel in the X-ray image depends on the amount of paint and primer present. Some artifacts, such as wood stretchers or nails, may also be visible. Figure~\ref{fig:Taffeta}(b) includes samples of 1 cm-sided X-ray plates, illustrating that plain weave may exhibit different textures even with similar thread counts. Additionally, in some areas of the canvases, quite noisy or rotated samples may be encountered.

\subsection{Motivation and Contributions}
Extensive literature~\cite{Johnson2010, Maaten15, Simois18, Bejarano2023} exists on thread counting and generating density maps for different fabrics. When these maps match, it can be concluded that the pieces of cloth were cut from the same roll. However, there is no tool to conclude the similarity between fabrics if such a match is not found. This could be because the fabrics originate from the same loom but different rolls, or are simply being cut from areas within the same roll that do not share warp or weft areas. This scenario is particularly common in the case of small-sized paintings.

 In this study, we investigate the possibility of comparing canvases through the analysis of their fabrics when a thread density match is not observed.
We approach the automatic comparison of fabrics from two canvases by comparing random samples from the X-ray plate images. We introduce a siamese deep learning (DL) model \cite{Koch2015, Chicco2021} with contrastive learning \cite{Hadsell2006}. This model is trained on canvases from the Museo Nacional del Prado (MNP), primarily from the 17th and 18th centuries. Random samples from canvases are compared, and the model parameters are adjusted to predict whether the samples originate from the same fabric or not. 

In the production phase, we propose to compare a set of random samples between pairs of canvases and record the model's predictions. These predictions are then analyzed to determine the similarity between the fabrics. Our aim is not only to identify differences in thread densities but also to differentiate fabrics with the same thread densities but distinct textures.

In summary, the main contributions of the paper are as follows:
\begin{itemize}
\item A new automatic tool to conclude the similarity between fabrics from paintings, not based on thread density map matching.
\item A novel Siamese DL model design and training approach to compare images from X-rays of canvases.
\item A method to provide an estimation of similarity from the set of predictions for pairs of samples from two fabrics.
\item Application to canvases from the MNP that sustains the starting hypothesis on the viability of comparing plain weave canvases, even in cases of similar thread counts.
\end{itemize}

\section{Related Works}

Image processing has been applied to the study of paintings \cite{Barni05} for canvas removal \cite{Deligiannis2017}, crack detection \cite{Cornelis13}, cradle removal \cite{Yin14}, chemical element analysis \cite{Yan2021} or style analysis \cite{ Li2004, Johnson08, Buchana2016} among others. Recently, algorithms in the machine learning field have been applied. In \cite{Rucoba22}, crack detection is revisited by applying K-SVD, and in \cite{Sizyakin20}, convolutional neural networks (CNN) were applied. CNN was also applied to the automatic classification of paintings \cite{Roberto2020}. In \cite{Pu2020}, auto-encoders were used for image separation. In \cite{Zou21}, DL was applied for virtual restoration of colored paintings. Style or authorship is predicted for isolated masterpieces after training a neural network-based model, as discussed in~\cite{Castellano2020} and the references therein. 

When using fabric analysis for forensic purposes, the theoretical results on frequency analysis applied to fabric analysis~\cite{Escofet2001, Johnson2013, Simois18} have been widely exploited. A repetition pattern, such as the crossings between horizontal and vertical threads in plain weave, can be described as a quasi-periodic 2D function in both vertical and horizontal dimensions. The maximum values of this pattern are related to the thread densities of the plain weave, which are interesting features to analyze. Frequency analysis tools allow for an easy finding of these maxima. By applying this approach locally through the canvas, we get a map of densities for the warp and weft at every point. A match between these maps for two canvases might indicate the same authorship, dating, being part of a series, etc. 


The main advantage of this frequency analysis \cite{Johnson2013, Simois18} approach is its unsupervised nature, i.e., it does not require any previous labeling for training. Furthermore, it is usually robust in the presence of noise, such as the painting itself or artifacts in the X-ray. Another approach proposed to study thread densities in canvases is based on feature extraction and machine learning~\cite{Maaten15}. This approach works in the spatial domain to locate crossing points and then estimate the thread densities. The major drawback is the need to annotate part of the work at hand, i.e., an expert must indicate the crossing points for a given area of the painting to be analyzed. Recently, DL has been proposed to segment crossing points~\cite{Bejarano2022a, Bejarano2022b, Bejarano2023} with improved results: a U-Net~\cite{Unet15} based model is introduced to locate crossing points in a 1 cm-sided square input image. Then, signal processing is used to estimate the average distance between crossing points. In~\cite{Bejarano2023b}, a regression model was used instead, directly estimating the thread density for the given input image.

The main drawback of the previous studies is that we need the canvases to share threads in the roll, either in the weft or the warp. Large canvases usually use the whole width of the roll, but small paintings may not. In addition, two canvases may have fabrics from the same provider made on different looms but with the same techniques and materials. In these cases, the previous approach fails. Furthermore, we may find two cloths sewn in large paintings, and to find a match, we need to process every cloth independently. This requires a previous analysis by the curator. Besides, these analyses are time and computationally demanding. Hence, they are faced when the curator, after other analyses, conjectures that two masterpieces could share the fabric. In this work, we face the comparison of two fabrics without resorting to matching the thread densities of the canvases.

In the analysis of fabrics, some steps were taken towards the characterization of fabric type~\cite{Cornelis10}.  However, the study was limited to new (industrial) and raw fabrics with different thread counts, at least in the warp or weft. Still, the main results of fabric analysis focus on average thread counting or density estimation.  
To our knowledge, no previous work has focused on a practical automatic method to assess the similarity between fabrics of old paintings not based on the match of thread count maps.  


%

\section{Siamese Model}
\begin{figure}[htbp]
\centering
\includegraphics[width=7cm]{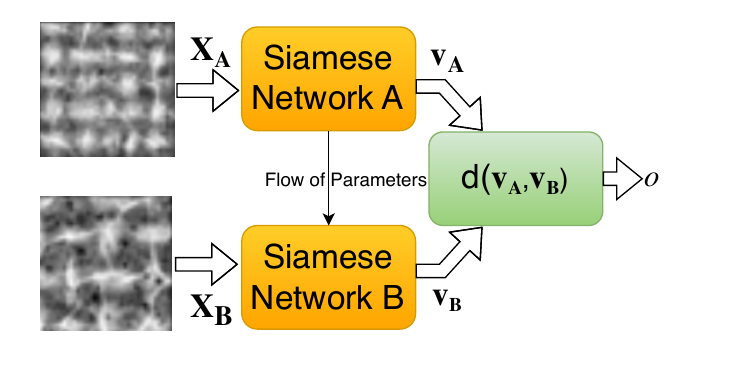}
\caption{Model of the Siamese network.\label{fig:Siamese}}
\end{figure}

The solution pipeline can be divided into training and production (or test) stages. In the training stage, we first gather the data, then develop a Siamese neural network (SNN)  model \cite{Bromley1993, Baldi1993, Koch2015} incorporating contrastive loss, and finally train the solution.
A brief description of the SNN used is presented in Figure~\ref{fig:Siamese} and consists of two input branches, where the inputs to each branch are a pair of images to be compared. Each image passes through the same deep learning architecture, i.e., Siamese Network A and Siamese Network B are two instances of the same architecture with the same parameters. The role of the Siamese network is to describe the fabric, providing some encoded features, $\vect{v}$. Then, the output of each branch is input to a block that compares them by using a distance measure, $\textrm{d}(\cdot,\cdot)$, to provide an estimation of similarity, $o\geq 0$. Low values for $o$ indicate similar fabrics. In production, given a pair of fabric images to compare, we generate a random crop from each image, which is then used as input to the model. This process is repeated $N$ times, and the results are finally analyzed collectively.

\subsection{Data Generation}
We selected a set of paintings and labeled them into classes. Paintings known to use fabrics from the same roll are included in the same class. Canvases from the Museo Nacional del Prado (MNP) by artists such as Rubens, Velázquez, Lorena, Swanevelt, Dughet, Poussin, Both, Lemaire, Ranc, Teniers, Miguel de Pret, and Ribera, among others, were included in the dataset. Paintings were used in only one subset: training, validation, or test. The number of paintings used in each group was 30, 15, and 12, respectively.  We try to include in each group representative samples of different types of fabrics with different thread densities. We group paintings known to share the same fabric into the same class, resulting in 18 and 11 groups of different fabrics for the training and validation data sets, respectively. 

All images of X-ray plates are preprocessed to enhance the fabric. The full description of the preprocessing can be found in \cite{Bejarano2023b} and encompasses a resampling to 200 pixels per cm, a mean and local standard deviation filtering of variable window size, and histogram equalization. Then we randomly select $M$ square patches of side 300 pixels, hereafter referred to as \textit{samples}. Inputs to the model are square patches of 100 pixels side, denoted by \textit{instances}. Instances are generated from samples by applying data augmentation as follows.
%
%
In the training and validation, we feed the model with 10 instances of side $100$ pixels from each of the $M$ 300 pixels sided samples, corresponding to the 4 corners and the central part plus their horizontal flips. 
Then, every instance can be randomly vertically flipped and/or rotated by $90^\textrm{o}$. Rotation is included as two paintings can share the same fabric but when framed they could have been rotated by $90^\textrm{o}$, and the training should consider this possibility.


\subsection{Siamese Network}

\begin{figure}[htbp]
\centering
\includegraphics[width=9cm]{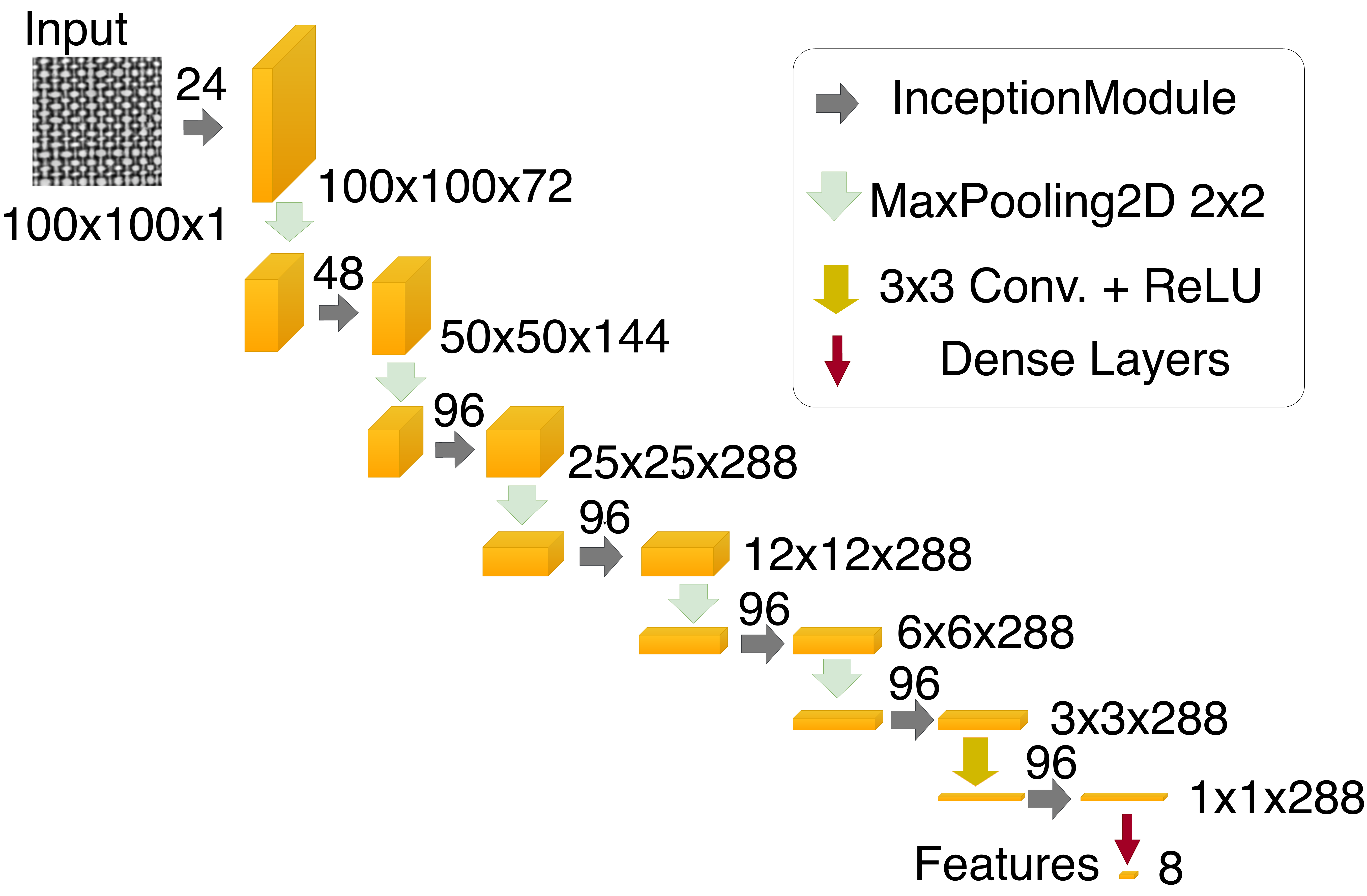}
\caption{Siamese network based on inception.}
\label{fig:SiameseMod}
\end{figure}

As Siamese network, we analyze three different models. On the one hand, two well-known architectures, the DenseNet121 (denoted below by DN121) \cite{Huang2017} and the VGG16 \cite{Simonyan2015}, on the other, we propose a model as follows. This model, hereafter denoted as the inception model \cite{Inception14}, borrows from \cite{Bejarano2023} the use of three different kernel sizes in the convolutional layers to cope with different thread densities of the fabrics. 
It has 7 layers, resembling an encoder network. The first 5 layers are CNN inception-based. Layer 6 is a CNN layer with ReLU; after flattening, the final layer is a fully connected (FC) network. In the first 5 layers, we reuse a block hereafter denoted by \textit{inception module} that comprises 3 convolutional layers with batch normalization in parallel of kernel sizes 3, 5, and 7 of $n$ features each. A ReLU activation function is applied to later stack the outputs into one vector, with $3n$ features. The FC network has two intermediate layers with $1024$ and $256$ neurons. 
In the diagram in Fig. \ref{fig:SiameseMod}, horizontal arrows stand for inception blocks, numbers indicating the number of filters, $n$. In contrast, vertical arrows in the CNN part indicate a 2D max-pooling, while in the dense network, they are FC layers. FC layers use ReLU as activation functions except for the output neuron, which uses a linear activation\footnote{Code describing this model will be provided upon acceptance.}. We performed a grid search \cite{Snoek12,Omalley19} to set the optimal value for the number of convolutional and dense layers, the number of filters in each layer, and the number of neurons in each dense layer.

\subsection{Contrastive Loss}
We compare the vectors generated at the output of the two branches of the SNN. We want to tune the weights to provide a 0 at the output if the two instances come from the same fabric and 1 otherwise. We propose to use a contrastive loss function \cite{Hadsell2006} as follows:
\begin{equation}
\mathcal{L}(\vc{v}_a, \vc{v}_b) = (1 - y) \cdot \left(\mathrm{d}(\vc{v}_a, \vc{v}_b)\right)^2 + y \cdot \left(\max\left(m - \mathrm{d}(\vc{v}_a, \vc{v}_b), 0\right)\right)^2
\end{equation}
where $y$ is the label, being 0 for similar and 1 otherwise, $\mathrm{d}(\vc{v}_a, \vc{v}_b)$ is the Euclidean distance between vectors $\vc{v}_a$ and $\vc{v}_b$, and $m$ is a margin value. The margin, along with the $\max$ function, allows us to prevent vectors already distant by $m$ or larger from being included in the loss.


\subsection{Training}

\begin{figure}[htp]
\centering
\begin{tikzpicture}

\begin{axis}[
legend cell align={left},
legend style={fill opacity=0.8, draw opacity=1, text opacity=1, draw=white!80!black},
tick align=outside,
tick pos=left,
x grid style={white!69.0196078431373!black},
xlabel={M},
xmajorgrids,
xmin=15, xmax=165,
xtick style={color=black},
y grid style={white!69.0196078431373!black},
ylabel={Loss},
ymajorgrids,
ymin=0.09, ymax=0.3,
ytick style={color=black},
ytick={0.075,0.1,0.125,0.15,0.175,0.2,0.225,0.25,0.275,0.3},
yticklabels={0.0,0.1,0.2,0.3,0.4,0.5,0.6,,,}
]
\addplot [semithick, blue, dashed, mark=diamond*, mark size=3, mark options={solid}]
table {%
20 0.532603941857815
40 0.235277819260955
80 0.106274438463151
160 0.101907472498715
};
\addlegendentry{DN121 Test}
\addplot [semithick, red, dashed, mark=*, mark size=3, mark options={solid}]
table {%
20 0.496336323022842
40 0.244559279084206
80 0.115969115495682
160 0.109972545597702
};
\addlegendentry{VGG16 Test}
\addplot [semithick, green!50!black, dashed, mark=+, mark size=3, mark options={solid}]
table {%
20 0.496689587831497
40 0.245014781877399
80 0.118757987581193
160 0.101264279987663
};
\addlegendentry{Inception Test}
\addplot [semithick, blue, mark=diamond*, mark size=3, mark options={solid}]
table {%
20 0.213963091373444
40 0.148777290061116
80 0.126876162923872
160 0.115211925469339
};
\addlegendentry{DN121 Validation}
\addplot [semithick, red, mark=*, mark size=3, mark options={solid}]
table {%
20 0.21000304594636
40 0.163954055681825
80 0.133365334384143
160 0.133619711920619
};
\addlegendentry{VGG16 Validation}
\addplot [semithick, green!50!black, mark=+, mark size=3, mark options={solid}]
table {%
20 0.17499863281846
40 0.142870907858014
80 0.112570536509156
160 0.10772752398625
};
\addlegendentry{Inception Validation}
\end{axis}

\end{tikzpicture}
\caption{Evolution of the training (solid) and test (dashed) losses for the proposed model ({\color{ForestGreen}{$+$}}) compared to the DenseNet121 ({\scriptsize{\color{blue}{$\blacklozenge$}}}) and the VGG16 ({\color{red}{$\circ$}}) architectures. The lowest loss out of 10 training with different weight initialization is depicted.}  
\label{fig:M}
\end{figure}
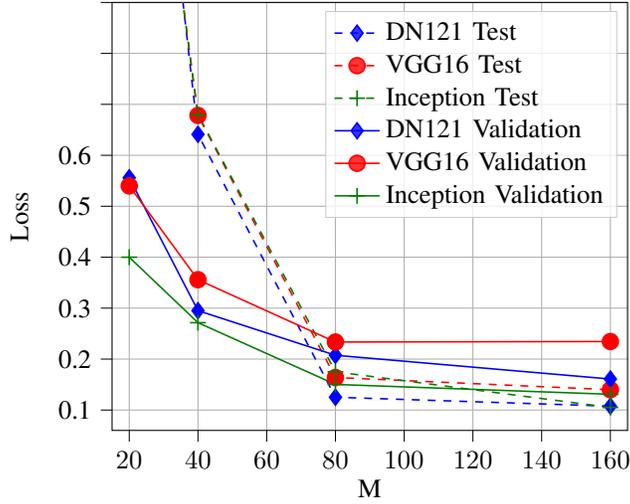

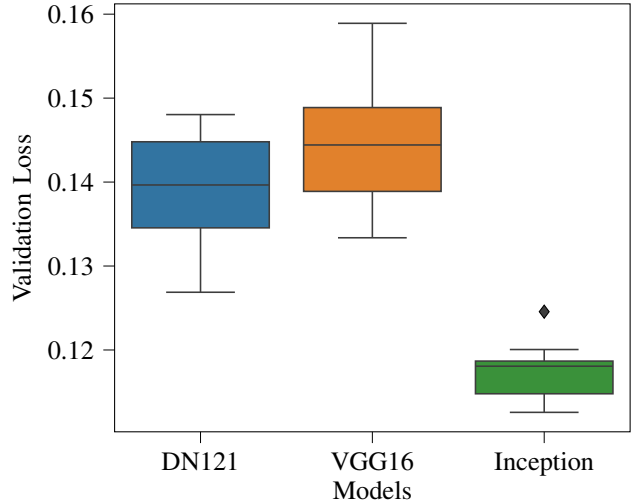
\begin{figure}[htp]
\centering
\begin{tikzpicture}

\definecolor{color0}{rgb}{0.194607843137255,0.453431372549019,0.632843137254902}
\definecolor{color1}{rgb}{0.881862745098039,0.505392156862745,0.173039215686275}
\definecolor{color2}{rgb}{0.229411764705882,0.570588235294118,0.229411764705882}

\begin{axis}[
tick align=outside,
tick pos=left,
x grid style={white!69.0196078431373!black},
xlabel={Models},
xmin=-0.5, xmax=2.5,
xtick style={color=black},
xtick={0,1,2},
xticklabels={DN121,VGG16,Inception},
y grid style={white!69.0196078431373!black},
ylabel={Validation Loss},
ymin=0.110253597432747, ymax=0.16122625711374,
ytick style={color=black},
ytick={0.11,0.12,0.13,0.14,0.15,0.16,0.17},
yticklabels={0.11,0.12,0.13,0.14,0.15,0.16,0.17}
]
\path [draw=white!23.921568627451!black, fill=color0, semithick]
(axis cs:-0.4,0.134547733142972)
--(axis cs:0.4,0.134547733142972)
--(axis cs:0.4,0.144799390947446)
--(axis cs:-0.4,0.144799390947446)
--(axis cs:-0.4,0.134547733142972)
--cycle;
\path [draw=white!23.921568627451!black, fill=color1, semithick]
(axis cs:0.6,0.138889873865992)
--(axis cs:1.4,0.138889873865992)
--(axis cs:1.4,0.148874707520008)
--(axis cs:0.6,0.148874707520008)
--(axis cs:0.6,0.138889873865992)
--cycle;
\path [draw=white!23.921568627451!black, fill=color2, semithick]
(axis cs:1.6,0.114784230990335)
--(axis cs:2.4,0.114784230990335)
--(axis cs:2.4,0.118677842151374)
--(axis cs:1.6,0.118677842151374)
--(axis cs:1.6,0.114784230990335)
--cycle;
\addplot [semithick, white!23.921568627451!black]
table {%
0 0.134547733142972
0 0.126876162923872
};
\addplot [semithick, white!23.921568627451!black]
table {%
0 0.144799390947446
0 0.148038679175079
};
\addplot [semithick, white!23.921568627451!black]
table {%
-0.2 0.126876162923872
0.2 0.126876162923872
};
\addplot [semithick, white!23.921568627451!black]
table {%
-0.2 0.148038679175079
0.2 0.148038679175079
};
\addplot [semithick, white!23.921568627451!black]
table {%
1 0.138889873865992
1 0.133365334384143
};
\addplot [semithick, white!23.921568627451!black]
table {%
1 0.148874707520008
1 0.158909318037331
};
\addplot [semithick, white!23.921568627451!black]
table {%
0.8 0.133365334384143
1.2 0.133365334384143
};
\addplot [semithick, white!23.921568627451!black]
table {%
0.8 0.158909318037331
1.2 0.158909318037331
};
\addplot [semithick, white!23.921568627451!black]
table {%
2 0.114784230990335
2 0.112570536509156
};
\addplot [semithick, white!23.921568627451!black]
table {%
2 0.118677842151374
2 0.120053691975772
};
\addplot [semithick, white!23.921568627451!black]
table {%
1.8 0.112570536509156
2.2 0.112570536509156
};
\addplot [semithick, white!23.921568627451!black]
table {%
1.8 0.120053691975772
2.2 0.120053691975772
};
\addplot [black, mark=diamond*, mark size=2.5, mark options={solid,fill=white!23.921568627451!black}, only marks]
table {%
2 0.124560390412807
};
\addplot [semithick, white!23.921568627451!black]
table {%
-0.4 0.139654610678554
0.4 0.139654610678554
};
\addplot [semithick, white!23.921568627451!black]
table {%
0.6 0.144419647473842
1.4 0.144419647473842
};
\addplot [semithick, white!23.921568627451!black]
table {%
1.6 0.11806766372174
2.4 0.11806766372174
};
\end{axis}

\end{tikzpicture}
\caption{Box diagram of validation loss for the validation set and 10 trainings for the proposed approach compared to models based on the DenseNet121 and the VGG16 architectures for $M=80$. Outliers are represented with diamonds.}  
\label{fig:Val}
\end{figure}

The model is presented with a pair of instances: if they come from the same fabric, i.e. belong to the same class, the label for the output is set to 0, and to 1 otherwise. 
In the training and validation, we need to feed the model with samples taken from different and same types of fabrics. 
Since it is harder to learn when samples correspond to the same class, in the training stage, pairs of instances from the same class are presented with probability $p=3/4$. We also need to define the number of samples used. We study the evolution of the loss along $M$, included in Fig. \ref{fig:M}. The loss for the validation (solid) and test (dashed) datasets is depicted for $M= 20, 40, 80, 160$ for the DenseNet121 ({\scriptsize{\color{blue}{$\blacklozenge$}}}), the VGG16 ({\color{red}{$\circ$}}), and the inception model ({\color{ForestGreen}{$+$}}) architectures. 
In DenseNet121 and VGG16, global average pooling and a dense layer were used as the final layers. Models were trained 10 times using different random weight initializations. In the figure, we retain the lowest value achieved. It can be observed that the proposed model has the lowest validation loss. 
%
%
In Fig. \ref{fig:Val} we include the box diagram for the validation losses of each architecture and the 10 trainings. It can be observed that the inception model exhibits a slightly better performance. The parameters for the training of the models were set as follows. In all models, the batch size was set to 256, early stopping was used with patience $20$, and the learning rate was set to $6\cdot10^{-5}$ and divided by $3$ every $25$ epochs. The DenseNet121 and VGG16 were trained with Adam, while the Inception model was trained with stochastic gradient descent with momentum $0.9$. 

%

\subsection{Production}\label{ssec:prod}


In production, comparing just a pair of instances is meaningless. As outlined above, we generate $N$ pairs of instances for the analysis. 
Let us denote the X-ray images of the canvases to be compared as $\ma{A}$ and $\ma{B}$. We first get a set of $N$ instances by randomly cropping $ 100$-pixel square regions. 
%
%
We feed the model in Fig. \ref{fig:Siamese}, once trained, with every pair, $\matr{X}_{A,i}, \matr{X}_{B,i}$, $i = 1, \ldots, N$, to obtain a vector of outputs $\vc{o}_{AB} = [o_{AB,1}, o_{AB,2}, \ldots, o_{AB,N}]$, $o_{AB,i} \geq 0$. 

To check for the $N$ pairs to decide on the similarity, we propose to compare the probability density functions (pdf's) of the outcomes, $\vc{o}_{AB}$, with the outcomes when comparing $\ma{A}$ to itself, $\vc{o}_{AA}$, used as baseline.
We generate another set $X'_{A,j}$, where the $j$th entry is randomly taken from the samples, and feed the model with pairs $\matr{X}_{A,i}, \matr{X}'_{A,i}$ to obtain $\vc{o}_{AA} = [o_{AA,1}, o_{AA,2}, \ldots, o_{AA,N}]$. 
Next, we estimate the pdf's for $\vc{o}_{AA}$, 
and $\vc{o}_{AB}$ as $p_{AA}(o)$, 
and $p_{AB}(o)$, respectively. We use a histogram analysis with $K$ bins with support from $0$ to $t$. Finally, we produce a measurement by using the symmetric Jensen-Shannon divergence (JSD) \cite{Lin1991}, also known as total divergence to the average or information radius \cite{Nielsen2019}:
\begin{align}
{j}_{A,B} &= \frac{1}{2}\left(\mathrm{KL}(p_{AA}(o) \|  \overline{p}_{AA,AB}(o)) \right.\notag\\  & \quad\quad\quad\quad\quad\quad \left.+ \mathrm{KL}(p_{AB}(o) \| \overline{p}_{AA,AB}(o))\right),\label{eq:JSD}
\end{align}
where we compute the Kullback-Leibler (KL) divergence between $p_{AA}(o)$ and $\overline{p}_{AA,AB}(o)$, and between $p_{AB}(o)$ and $\overline{p}_{AA,AB}(o)$, where  
%
 $\overline{p}_{AA,AB}(o) = (p_{AA}(o) + p_{AB}(o)) / 2$ is the average value of the pdf's. Compared to the Jeffreys' divergence the JSD is bounded and we avoid numerical issues related to empty events or bins.
%


Note that we compare $A$ and $B$ by measuring the JSD between the distribution of the outputs of the SNN with $A$ and $B$ as input, given by samples in $\vc{o}_{AB}$, and then $A$ and itself, given by $\vc{o}_{AA}$. If $A$ and $B$ were similar, we should obtain a similar pdf in both cases.
Besides, we could estimate ${j}_{B,A}$, which should provide a similar value to ${j}_{A,B}$ if $B$ and $A$ come from the same type of fabric. If not, as $p_{AA}(o)$ and $p_{BB}(o)$ should differ, these values would be different. We propose to estimate the symmetric indicator 
\begin{equation}
{s}_{B,A}=s_{A,B} = \max\left({{j}_{A,B},{j}_{B,A}}\right).\label{eq:SJSD}
\end{equation}
In Alg. \ref{alg:JSD} we describe the steps to estimate this metric from the images using the SNN. We use a threshold, $u$, to limit the maximum values for better representation.

\begin{algorithm}
\caption{Symmetric JSD Estimation}\label{alg:JSD}
\textbf{Input}: Grayscale images $\matr{A}$ and $\matr{B}$, number of samples $N$, threshold $u$, number of bins $K$ and support $t$.
\begin{algorithmic}[1]
\STATE Get $N$ random instances from $\matr{A}$ and $\matr{B}$ as $\matr{X}_{A,i}$ and $\matr{X}_{i,B}$, $i=1,...,N$.  
\STATE Obtain the output of the SNN, $o_{A,B}$, for input every pair in the sample sets, $\matr{X}_{A,i}$ and $\matr{X}_{B,i}$.
\STATE Get another set of $N$ random instances from $\matr{A}$ as $\matr{X}'_{A,i}$.  
\STATE Obtain the output of the SNN, $o_{A,A}$, for inputs every pair in the sample sets, $\matr{X}_{A,i}$ and $\matr{X}'_{A,i}$.
\STATE Get another set of $N$ random instances from $\matr{B}$ as $\matr{X}'_{B,i}$.
\STATE Estimate the pdf of  $o_{A,B}$ as vector $q_{AB}$ by using histogram with $K$ bins in the range $(0,t)$.
\STATE Estimate the pdf of  $o_{A,A}$ as vector $q_{AA}$ by using histogram with $K$ bins in the range $(0,t)$.
\STATE Compute the average estimation of pdf's as $q_{AA,AB}=(q_{AA}+q_{AB})/2$ and $q_{BB,AB}=(q_{BB}+q_{AB})/2$.
\STATE Compute the estimation of the KL divergences from the pdf's estimations as $l_{AA}=\textrm{KL}(q_{AA}||\overline{q}_{AA,AB})$, $l_{BB}=\textrm{KL}(q_{BB}||\overline{q}_{BB,AB})$ and $l_{AB}=\textrm{KL}(q_{AB}||\overline{q}_{AB,AB})$.
\STATE Compute $\hat{j}_{A,B}=(l_{AA}+l_{AB})/2$ and $\hat{j}_{B,A}=(l_{BB}+l_{AB})/2$ as the estimation in (\ref{eq:JSD}) for ${j}_{A,B}$ and ${j}_{A,B}$, respectively.
\STATE Compute the estimation of the symmetric JSD as $\hat{s}_{A,B}=\max(\hat{j}_{A,B}, \hat{j}_{B,A})$.
\\
\textbf{Output:} $\min(\hat{s}_{A,B},u)$.
\end{algorithmic}
\end{algorithm} 



%
%
%
%

\section{Analysis of Canvases}\label{sec:Test}
We apply the proposed method to analyze the test set of canvases at the MNP.  The goal is to automatically highlight possible similarities between masterpieces by focusing solely on the fabrics. We predict which canvases, out of the 12 in the test subset, have similar fabrics. 
Every canvas is compared to the rest to assess the fabrics' similarity. We will work with X-ray plate images of the canvases. 

\begin{figure}[t] 
    \centering
    \subfloat[]{
        \includegraphics[width=0.2\linewidth]{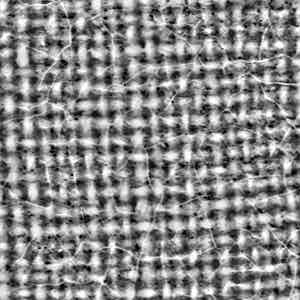}
    }
    \subfloat[]{
        \includegraphics[width=0.2\linewidth]{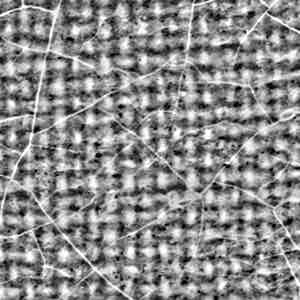}
    }
    \subfloat[]{
        \includegraphics[width=0.2\linewidth]{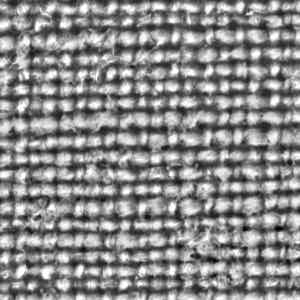}
    }
    \caption{Examples of samples from X-ray plates of (a) MPret7905 \cite{P007905}, (b) MPret7906 \cite{P007906} and (c) RubensImac \cite{P001627}.}
    \label{fig:MPretInm}
\end{figure}

The proposed case study involves challenging analyses. On the one hand, we have two paintings by Miguel de Pret, ``Two Bunches of Grapes" \cite{P007906} and ``Two Bunches of Grapes with a Fly" \cite{P007905}, denoted hereafter by MPret7906 and MPret7905, respectively. In \cite{Pret13}, it was concluded that both canvases share the same fabric, with high probability. However, as further developed in \cite{Simois18}, both canvases have a difference or more than one thread per cm in the densities of vertical threads, see the estimated thread densities in Tab. \ref{tab:densities} for horizontal threads. We also include in the test set a canvas by Rubens, ``The Immaculate Conception" \cite{P001627}, referred to as RubensInm. The fabric of this canvas has thread density values closer to those of MPret7906 than MPret7905, but its fabric is of another type, as discussed in \cite{Simois18}. In Fig. \ref{fig:MPretInm}, samples for three of these canvases are included. In Fig. \ref{fig:MPretInm}.(c), the sample from RubensImc, exhibits horizontal threads much wider than vertical ones, while in Fig. \ref{fig:MPretInm}.(a) and Fig. \ref{fig:MPretInm}.(b) widths of horizontal and vertical threads are similar.

\begin{table*}[!t]
\caption{Thread densities of horizontal and vertical threads, in threads per cm, for the fabrics of the canvases in the test dataset. \label{tab:densities}}
    \centering
    \begin{tabular}{lcccc}
        \hline
        \textbf{Authorship} & \textbf{Date} & \textbf{Canvas} & \textbf{Horiz.} & \textbf{Vert.} \\
        \hline
        Miguel de Pret & 1630-1644 & MPret7905& 12.12 & 13.68 \\
        Miguel de Pret & 1630-1644 & MPret7906 & 10.16 & 12.90 \\
        Juan Bautista Mart{\'i}nez del Mazo & 1665-66 & MMazo & 6.65 & 8.98 \\
        Bartolom{\'e} Esteban-Murillo& 1660-65&Muri998 & 10.74 & 11.92 \\
        Bartolom{\'e} Esteban-Murillo& 1660-65&Muri999 & 11.33 & 11.92 \\
        Bartolom{\'e} Esteban-Murillo& 1660-65&Muri1000 & 10.93 & 12.52 \\
        Luis Paret y Alc{\'a}zar& 1772-1773& Paret2991& 20.71 & 21.10 \\ 
        Luis Paret y Alc{\'a}zar& 1777& Paret7701 & 21.98 & 20.71 \\ 
        Luis Paret y Alc{\'a}zar& 1776 &Paret60482 & 21.88 & 19.14 \\
        Peter Paul Rubens & 1628-29 &RubensInm & 10.75 & 13.09 \\
       Diego Rodr{\'\i}guez de Silva y Vel{\'a}zquez& 1632& Velaz1195 & 14.85 & 18.36 \\
       Diego Rodr{\'\i}guez de Silva y Vel{\'a}zquez& 1632& Velaz1196 & 14.46 & 17.97 \\
        \hline
    \end{tabular}
\end{table*}

\begin{figure}[t] 
    \centering
    \subfloat[]{
        \includegraphics[width=0.2\linewidth]{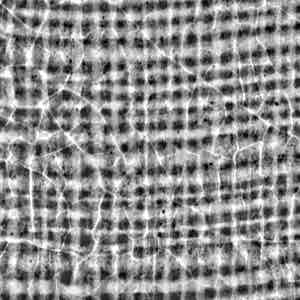}
    }
    \subfloat[]{
        \includegraphics[width=0.2\linewidth]{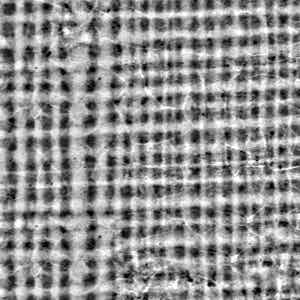}
    }
    \subfloat[]{
        \includegraphics[width=0.2\linewidth]{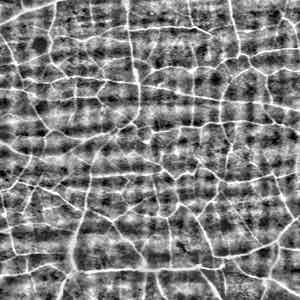}
    }
    \caption{Examples of samples from X-ray plates of (a) Muri998 \cite{P000998}, (b) Muri999 \cite{P000999} and (c) Muri1000 \cite{P001000}.}
    \label{fig:Murillo}
\end{figure}

Then we include three canvases by Murillo: ``The Prodigal Son taking leave of his Home" \cite{P000998}, ``The Prodigal Son squandering his Inheritance" \cite{P000999}, and ``The Prodigal Son among the Pigs" \cite{P001000}, in the following denoted by Muri998, Muri999 and Muri1000, respectively.  
They correspond to the same series. It has been proved \cite{Bejarano2023} that Muri998 and Muri999 share the same fabric. Since the three canvases belong to the same series of paintings devoted to the Prodigal Son, we conjecture that Muri1000 shares the type of fabric with Muri998 and Muri999. We include a sample from each canvas in Fig. \ref{fig:Murillo}.  Note that the sample from Muri1000 in Fig. \ref{fig:Murillo}.c has more cracks than the other two paintings. We observed that samples from this canvas were of poorer quality than those obtained from the other two. Note that these canvases have densities close to those of the canvases by Miguel de Pret and Rubens.

The canvas by Martínez del Mazo ``Margarita de Austria" \cite{P000888}, denoted as MMazo hereafter, is also included. This canvas has not been reported to share its fabrics with any other in the test dataset. Density values, see Tab. \ref{tab:densities}, are quite different compared to the other canvases.

We added to the test dataset the canvases ``Diego Corral y Arrellano" \cite{P001195} and ``Antonia de Ipe{\~n}arrieta y Galdós and her son, Luis" \cite{P001196}, denoted by Velaz1195 and Velaz1196, respectively.  In these two paintings, Velázquez portrays a husband and his wife, and his son. Both canvases use the same fabric as there is a perfect match between patterns of horizontal thread densities, shown in \cite{Bejarano2023}. These canvases have much higher densities compared to the previous ones. 

\begin{figure}[t!] 
    \centering
    \subfloat[]{
        \includegraphics[width=0.2\linewidth]{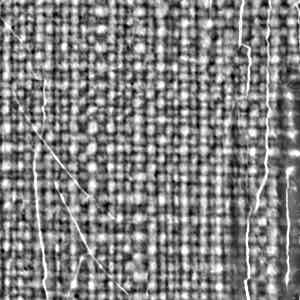}
    }
    \subfloat[]{
        \includegraphics[width=0.2\linewidth]{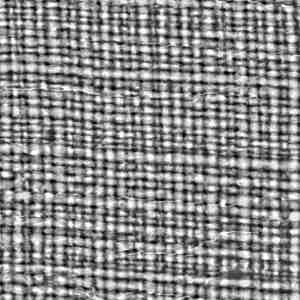}
    }
    \subfloat[]{
        \includegraphics[width=0.2\linewidth]{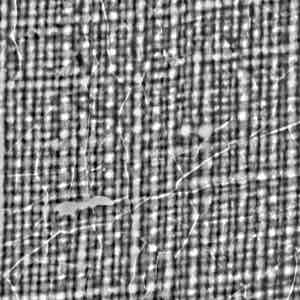}
    }
    \caption{Examples of samples from X-ray plates of (a) Paret2991 \cite{P002991}, (b)  Paret60482 and (c) Paret7701 \cite{P007701}.}
    \label{fig:Paret}
\end{figure}

Finally, we have three canvases by Luis Paret y Alc{\'a}zar, ``Boudoir Scen" \cite{P002991},  ``Self-portrait" and ``Self-portrait of the artist in his studio" \cite{P007701}, referred to as Paret2991, Paret60482, and Paret7701, respectively. These canvases have high densities of values, similar to those for the vertical threads of the canvas by Velázquez, see Tab. \ref{tab:densities}. In Fig. \ref{fig:Paret}, the samples of the three canvases exhibit a similar texture. Besides, they have close thread densities. We conjecture they share the same type of fabric. Note that, to illustrate the performance of the approach, we will not exploit the available dating, a century after the other ones, to discriminate between fabrics. 



In brief, we expect the analysis to associate the canvases into the following groups: i) MPret7905 and MPret7906, ii) Muri998, Muri999 and Muri1000, iii) Paret2991, Paret60482, and Paret7701, and iv) Velaz1195 and Velaz1196. Fabrics from MMazo and RubensImn should not be paired with any other one. However, solutions could focus on thread densities, failing to pay attention to texture. Given Tab. \ref{tab:densities}, the algorithm will find it hard to discriminate within the groups a) MPret7905, MPret7906, Muri998, Muri999, Muri1000, and RubensInc, and b) Paret2991, Paret60482, and Paret7701, Velaz1195, and Velaz1196.

%
%

We report in Fig. \ref{fig:S1}-\ref{fig:S4} the results of the analysis. 
We include a plot where the estimated similarity metric for every pair of test images of the fabrics, $(\matr{I},\matr{J}$), is represented in a grayscale grid image. 
We represent the values of the estimation of matrix $\matr{S}$, whose entry $\matr{S}_{i,j}$ is the indicator $s_{I,J}$ in (\ref{eq:SJSD}), for images at position $i$ and $j$. As developed in Section \ref{ssec:prod} we estimate the $s_{I,J}$ index for every pair. To estimate (\ref{eq:JSD}) we used Alg. \ref{alg:JSD} with $N=1000$, $u=0.03$, $K=50$ and $s=2.5$. These values were set through extensive experimentation. 
%
Recall that lower values, near 0, indicate high similarity. In the figure, these lower values are represented with darker colors. Note that on the diagonal we have black entries, as we are comparing samples from the same canvas. Out-of-diagonal, positions $i,j$, $i\neq j$, with low (dark) values indicate a match between fabrics of the corresponding canvases, $I$ and $J$.  Since $\matr{S}$ is symmetric, we will have the same value for $\matr{S}_{i,j}$ and $\matr{S}_{j,i}$. We set all values above $u=0.03$ to 1, displayed in white in Fig. \ref{fig:S1}-\ref{fig:S4}. High values indicate that the corresponding pairs of canvases are not good candidates for further studies of similitude.

%
%

\newcommand{\myimage}{gridN10S80DN121u3} 

\begin{figure}[!t] 
    \centering

\begin{tikzpicture}

\begin{axis}[
width=.35\textwidth, 
height=.35\textwidth,
tick align=outside,
tick pos=left,
x grid style={white!69.0196078431373!black},
xmin=-0.5, xmax=11.5,
xtick style={color=black},
xtick={0,1,2,3,4,5,6,7,8,9,10,11},
xticklabel style = {rotate=90.0},
xticklabels={MMazo,MPret7905,MPret7906,Muri1000,Muri998,Muri999,Paret2991,Paret60482,Paret7701,RubensImn,Velaz1195,Velaz1196},
y dir=reverse,
y grid style={white!69.0196078431373!black},
ymin=-0.5, ymax=11.5,
ytick style={color=black},
ytick={0,1,2,3,4,5,6,7,8,9,10,11},
yticklabels={MMazo,MPret7905,MPret7906,Muri1000,Muri998,Muri999,Paret2991,Paret60482,Paret7701,RubensImn,Velazq1195,Velazq1196}
]
\addplot graphics [includegraphics cmd=\pgfimage,xmin=-0.5, xmax=11.5, ymin=11.5, ymax=-0.5] {\myimage};
\end{axis}

\end{tikzpicture}
    \caption{Representation of matrix $\matr{S}$ for the DenseNet121 model and $M=80$. The darker, the lower the value, and the more similar the fabrics are.\label{fig:S1}}
\end{figure}
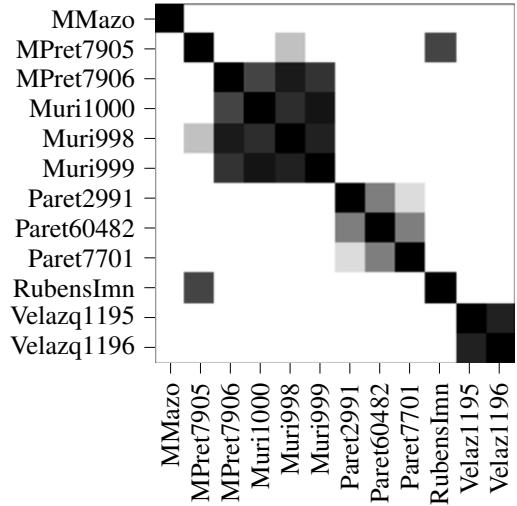

\renewcommand{\myimage}{gridN10S80Ownu3} 

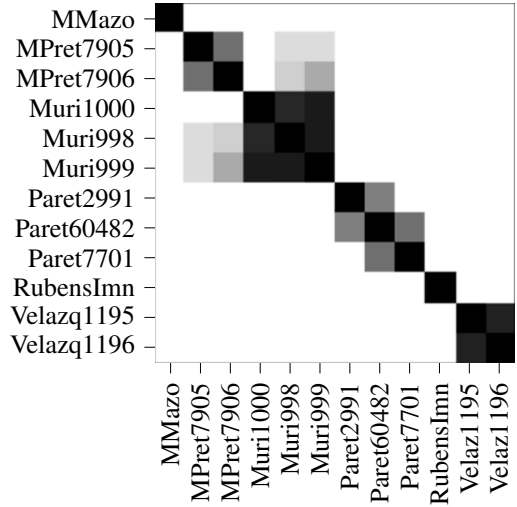
\begin{figure}[!t] 
    \centering

\begin{tikzpicture}

\begin{axis}[
width=.35\textwidth, 
height=.35\textwidth,
tick align=outside,
tick pos=left,
x grid style={white!69.0196078431373!black},
xmin=-0.5, xmax=11.5,
xtick style={color=black},
xtick={0,1,2,3,4,5,6,7,8,9,10,11},
xticklabel style = {rotate=90.0},
xticklabels={MMazo,MPret7905,MPret7906,Muri1000,Muri998,Muri999,Paret2991,Paret60482,Paret7701,RubensImn,Velaz1195,Velaz1196},
y dir=reverse,
y grid style={white!69.0196078431373!black},
ymin=-0.5, ymax=11.5,
ytick style={color=black},
ytick={0,1,2,3,4,5,6,7,8,9,10,11},
yticklabels={MMazo,MPret7905,MPret7906,Muri1000,Muri998,Muri999,Paret2991,Paret60482,Paret7701,RubensImn,Velazq1195,Velazq1196}
]
\addplot graphics [includegraphics cmd=\pgfimage,xmin=-0.5, xmax=11.5, ymin=11.5, ymax=-0.5] {\myimage};
\end{axis}

\end{tikzpicture}
    \caption{Representation of matrix $\matr{S}$ for the inception model and $M=80$. The darker, the lower the value, and the more similar the fabrics are.\label{fig:S2}}
\end{figure}


    \renewcommand{\myimage}{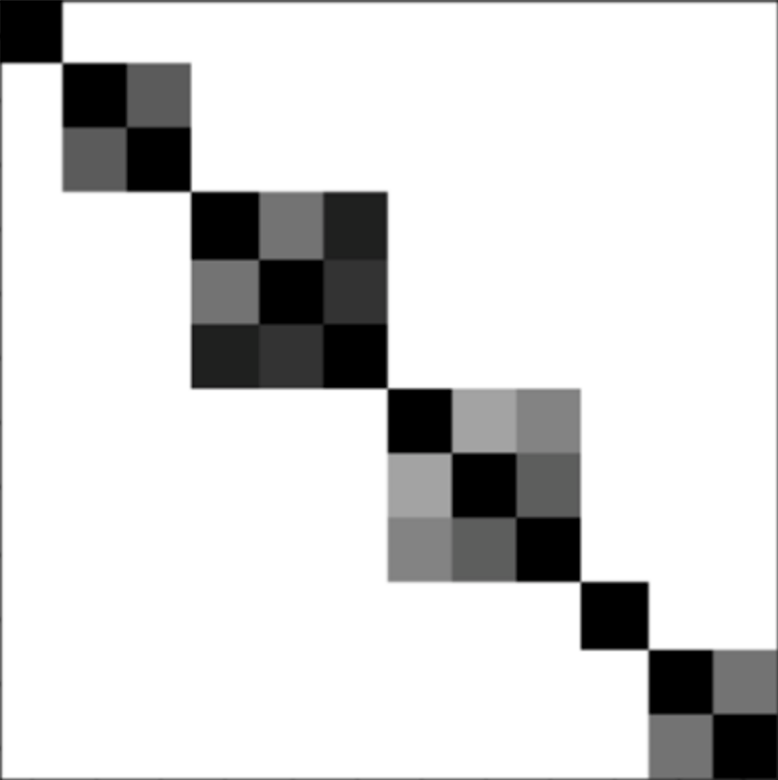}
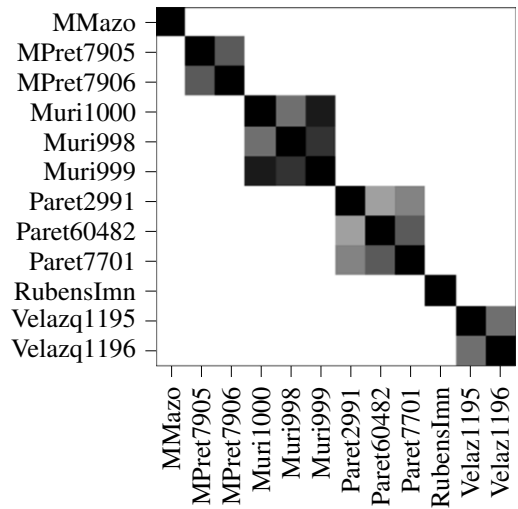
\begin{figure}[!h] 
    \centering

\begin{tikzpicture}

\begin{axis}[
width=.35\textwidth, 
height=.35\textwidth,
tick align=outside,
tick pos=left,
x grid style={white!69.0196078431373!black},
xmin=-0.5, xmax=11.5,
xtick style={color=black},
xtick={0,1,2,3,4,5,6,7,8,9,10,11},
xticklabel style = {rotate=90.0},
xticklabels={MMazo,MPret7905,MPret7906,Muri1000,Muri998,Muri999,Paret2991,Paret60482,Paret7701,RubensImn,Velaz1195,Velaz1196},
y dir=reverse,
y grid style={white!69.0196078431373!black},
ymin=-0.5, ymax=11.5,
ytick style={color=black},
ytick={0,1,2,3,4,5,6,7,8,9,10,11},
yticklabels={MMazo,MPret7905,MPret7906,Muri1000,Muri998,Muri999,Paret2991,Paret60482,Paret7701,RubensImn,Velazq1195,Velazq1196}
]
\addplot graphics [includegraphics cmd=\pgfimage,xmin=-0.5, xmax=11.5, ymin=11.5, ymax=-0.5] {\myimage};
\end{axis}

\end{tikzpicture}
    \caption{Representation of matrix $\matr{S}$ for the DenseNet121 model and $M=160$. The darker, the lower the value, and the more similar the fabrics are.\label{fig:S3}}
\end{figure}

\renewcommand{\myimage}{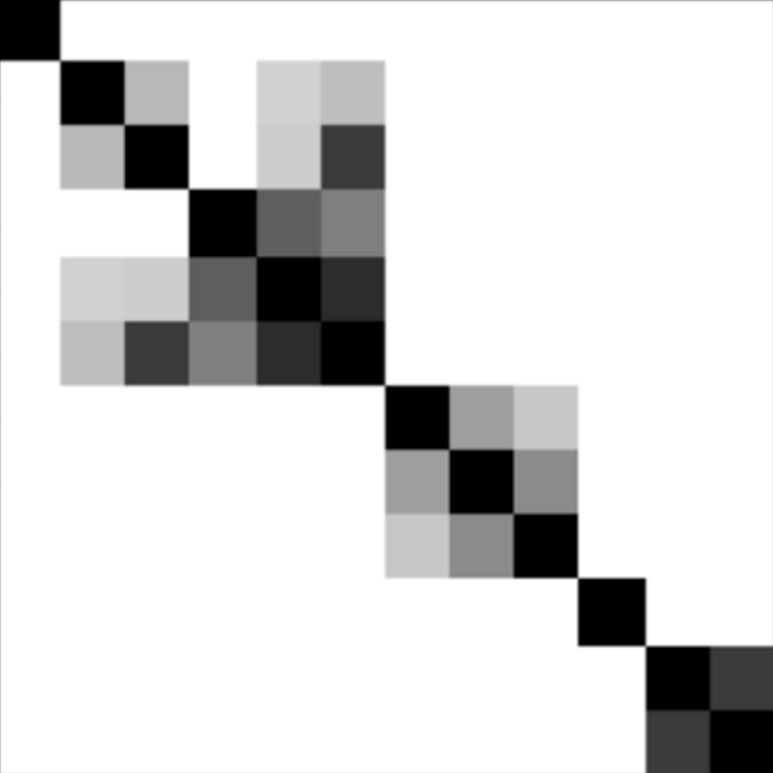}
\begin{figure}[!h] 
    \centering

\begin{tikzpicture}

\begin{axis}[
width=.35\textwidth, 
height=.35\textwidth,
tick align=outside,
tick pos=left,
x grid style={white!69.0196078431373!black},
xmin=-0.5, xmax=11.5,
xtick style={color=black},
xtick={0,1,2,3,4,5,6,7,8,9,10,11},
xticklabel style = {rotate=90.0},
xticklabels={MMazo,MPret7905,MPret7906,Muri1000,Muri998,Muri999,Paret2991,Paret60482,Paret7701,RubensImn,Velaz1195,Velaz1196},
y dir=reverse,
y grid style={white!69.0196078431373!black},
ymin=-0.5, ymax=11.5,
ytick style={color=black},
ytick={0,1,2,3,4,5,6,7,8,9,10,11},
yticklabels={MMazo,MPret7905,MPret7906,Muri1000,Muri998,Muri999,Paret2991,Paret60482,Paret7701,RubensImn,Velazq1195,Velazq1196}
]
\addplot graphics [includegraphics cmd=\pgfimage,xmin=-0.5, xmax=11.5, ymin=11.5, ymax=-0.5] {\myimage};
\end{axis}

\end{tikzpicture}
    \caption{Representation of matrix $\matr{S}$ for the inception model and $M=160$. The darker, the lower the value, and the more similar the fabrics are.\label{fig:S4}}
\end{figure}
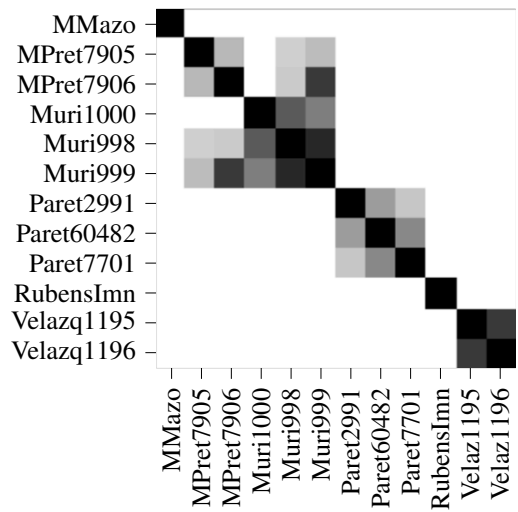

We depict the test results obtained for the architectures analyzed. For every model, we select the parameters providing the lowest validation loss, reported in Fig. \ref{fig:M}, 
%
for $M=80$ and $M=160$. These are the ones of the proposed inception model and DN121, respectively. For the sake of comparison, we analyze both models, for $M=80$ and $M=160$.
It can be observed in Fig. \ref{fig:S1} that for $M=80$ DenseNet121 fails to discriminate between RubensInm and MPret7905. As discussed above, although they exhibit close densities, their textures differ. Besides, it cannot differentiate the fabric of MPret7906 from that of Murillo. Furthermore, it does not find a correspondence between the fabrics of MPret7905 and Mpret7906. 
%
In the results in Fig. \ref{fig:S2} for the proposed inception model, this match is found, although a slight connection is indicated between Muri998-9 and MPret7905-6. Besides, it has difficulties pairing Paret2991 and Paret7701. This match can be inferred through the correspondence by pairs: since Paret2991 and Paret60482 are related, and so are Paret60482 and Paret7701, then Paret2991 and Paret7701 should share the same type of fabric. However, the proposed method should be more consistent, reporting a darker value for $\matr{S}$ at the position corresponding to the pair Paret2991-Paret7701.

%
%
%
%

As we increase the number of samples $M$ from $80$ to $160$ we observe in the results of the inception model, Fig. \ref{fig:S4}, that the masterpieces by Paret are all related, but it is unable to distinguish between the paintings by Murillo and Miguel de Pret. On the contrary, for $M=160$ the DN121 model, see Fig. \ref{fig:S3}, provides the expected outcome where paintings by authors are correctly grouped, and not related to the canvases by others. Hence, while for a low number of samples, the inception model provides better results, for a large enough number for $M$ the DenseNet121 model provides the expected outcome. The results observed for VGG16 were poor in both cases.

\section{Conclusions}
In this work, we propose a novel paradigm for canvas forensic studies based on the analysis of their fabrics. Previous works focus on given pairs of canvases to find matches of thread densities. Images of fabrics, usually of X-ray plates, are spatially processed to determine the thread density at any point of the canvas. Then the method searches for a match between the obtained patterns for both canvases. This process is quite useful, but it requires the fabrics of the canvas to have been cut from the roll, either aligned horizontally or vertically. In other words, they need to share the horizontal or vertical threads. This is usually the case for large canvases, as the whole roll width is used for the canvas. However, this may not be the case for smaller paintings. Furthermore, this analysis is time and computationally demanding, and it is usually accomplished when the curator, after analyzing a set of masterpieces, conjectures that a pair or a set of canvases could share the same type of fabrics. 

The proposed approach avoids this matching of density patterns. It just determines if a pair of canvases shares the same type of fabric based on random samples. This could be used as a first step to analyze sets of canvases and suggest matchings between fabrics. This would be quite useful as a large set can be analyzed with little assistance from the curator. A second relevant application of the proposal is to provide a measure of similitude between the fabrics of two canvases where density matching could not be achieved. 

The proposal is built upon recent advances in DL models. In these approaches, a large set of samples is needed while we have a limited set of images to train it. Besides, images of canvases have different sizes and dimensions, which pose difficulties in the design of the input layers of the DL models. We overcome these issues by comparing pairs of patches of the images of the fabrics. The trained model provides a measure of similarity given two patches from a pair of canvases. In production, we propose to compare a wide set of patches to statistically analyze the difference. We resort to the Janson-Shannon divergence. 

The results included prove that this new paradigm for fabric analysis is useful. The DL model focuses not just on the thread densities but also on the texture of the fabrics. It could be a milestone in the processing of fabrics at museums. Applied to a full museum collection or an author catalog, it would propose to the curator unknown associations between masterpieces to be further analyzed. 
Future research could be directed, among others, to 1) improve the DL models, 2) study preprocessing stages for feature extraction, 3) increase the dataset, and 4) improve the statistical analysis in production. 


%
%

%




 \bibliographystyle{IEEEtran}

\bibliography{art, DNN, Ours, Paintings, sigproc, Prado}
\end{document}